\documentclass[letterpaper, 10 pt, conference]{ieeeconf}  

\IEEEoverridecommandlockouts                              

\usepackage{xspace} 
\usepackage{amsmath}
\usepackage{amssymb}
\usepackage{booktabs}
\usepackage{multirow}
\usepackage{colortbl}
\usepackage{array}
\usepackage{xcolor}
\usepackage{graphicx}
\usepackage{eso-pic}

\usepackage[caption=false]{subfig}

\makeatletter
\let\NAT@parse\undefined
\makeatother
\usepackage[numbers,sort]{natbib}

\usepackage[colorlinks=true, citecolor=blue, linkcolor=blue, urlcolor=blue]{hyperref}

\AddToShipoutPictureFG*{%
  \put(0,760){%
    \makebox[\paperwidth]{%
      \parbox{0.8\paperwidth}{%
        \centering
        \scriptsize
        \textcopyright\ 2026 IEEE. Personal use of this material is permitted.
        Permission from IEEE must be obtained for all other uses, in any current
        or future media, including reprinting/republishing this material for
        advertising or promotional purposes, creating new collective works,
        for resale or redistribution to servers or lists, or reuse of any
        copyrighted component of this work in other works.
      }%
    }%
  }%
}

\title{\LARGE \bf
PRISM: 
Privileged 
\underline{Pr}obabil\underline{i}stic 
Latent 
\underline{S}upervision 
for End-to-End Autonomous Driving 
\underline{M}otion Planning
}

\vspace{-0.2cm}

\author{
    Volodymyr Havrylov$^{1,2}$, Faris Janjoš$^{2}$, Andreas Look$^{2,3}$, Jürgen Mathes$^{2}$, and Andreas Geiger$^{1,4}$ \\[0.1cm]
    {\small $^{1}$University of Tübingen \quad $^{2}$Bosch Center for Artificial Intelligence \quad $^{3}$Coburg University \quad $^{4}$Tübingen AI Center}
    \vspace{-0.2cm} 
}

\begin{document}

\maketitle
\thispagestyle{empty}
\pagestyle{empty}

\begin{abstract} 
End-to-end autonomous driving (E2E AD) systems integrate perception, prediction, and planning into a single differentiable architecture. While these models show great promise, their standard training often relies on output-only supervision, which can lead to weak gradients for the hidden layers of increasingly complex models. Recent works have integrated vision-language model (VLM) supervision for latent features to address this, yielding substantial empirical gains, yet leaving the underlying theoretical mechanisms poorly understood. Our investigation into this methodology reveals that the resulting performance gains stem not from VLM reasoning capabilities, as previously assumed, but rather from the latent connections forged between the E2E AD model and ground-truth (GT) data during training. Building on this insight, we propose a probabilistic deep supervision framework that regularizes intermediate latent representations directly from GT data. By treating model latents as reparameterizable distributions, we optimize the architecture via the Evidence Lower Bound (ELBO). Our evaluations conducted on the nuScenes dataset demonstrate that supervising trajectory-related latents with future GT paths consistently improves planning performance. Using identical training data and E2E architectures, our method achieves an 8\% reduction in planning L2 error and a 3\% decrease in collision rates compared to competitive vectorized baselines, all while incurring negligible computational overhead. The code is released at \url{link-available-soon}.
\end{abstract}

\section{INTRODUCTION}

Autonomous driving (AD) represents a transformative machine learning application that aims to fundamentally reshape global transportation by mitigating the human error responsible for the vast majority of traffic accidents \cite{nhtsa2015critical}. Simultaneously, it offers substantial economic and environmental benefits through optimized logistics networks that improve cost efficiency and reduce carbon footprints \cite{DBLP:journals/ejtl/LeeDHZ24, Gjere2023AutTruck}.

\begin{figure}[!t]
  \centering
  \includegraphics[width=1.0\columnwidth]{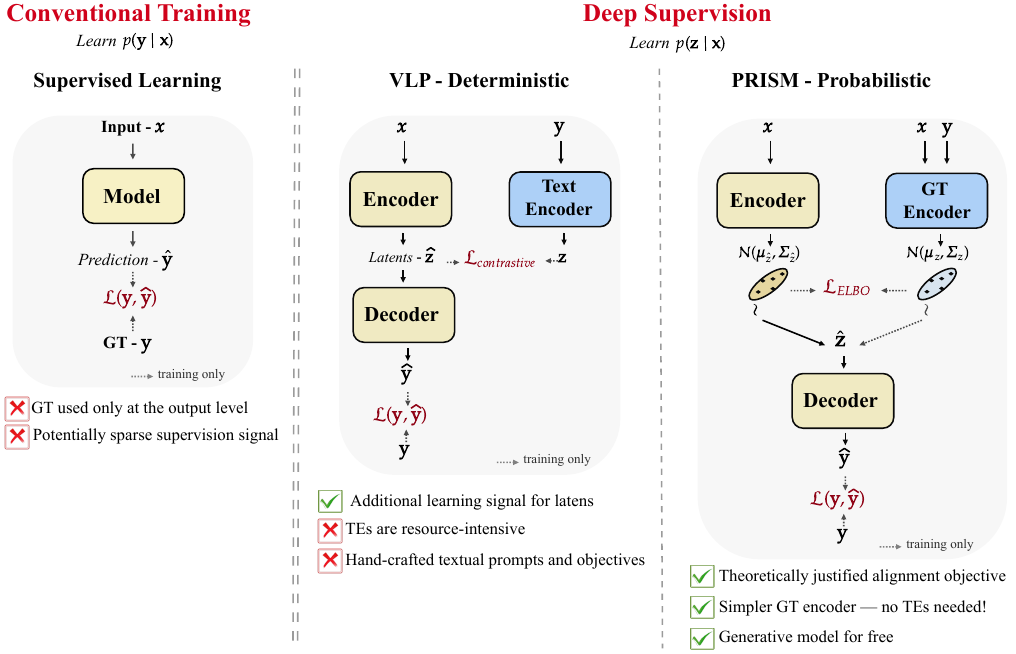}
  \caption{\textbf{High-level comparison of conventional supervised (left) and deep supervision (middle and right) training paradigms.} Our proposed deep supervision approach PRISM captures the interplay between a context $x$ and ground-truth future $y$ into a probabilistic representation without requiring expensive text encoders.
  }
  \label{fig:teaser}
\end{figure} 

Over the past decade, AD research has pivoted from traditional modular pipelines, which decompose the task into perception, prediction, and planning \cite{DICKMANNS1987221, urmson2008autonomous, yurtsever2020survey}, to end-to-end (E2E) learning approaches~\cite{DBLP:conf/nips/Pomerleau88, DBLP:conf/cvpr/HuYCLSZCDLWLJLD23, DBLP:conf/iccv/JiangCXLCZZ0HW23, DBLP:journals/corr/abs-2405-19620, DBLP:journals/corr/abs-2010-08776}. Modular systems, while interpretable, are notoriously susceptible to error propagation across modules. E2E models address this by unifying the stack into a single differentiable network, aiming to capture the full complexity of driving with joint optimization of all components toward the final objective.

While architectural innovations have been extensively studied, training paradigms for E2E AD models remain relatively underexplored. Most existing approaches rely on supervised learning with large-scale datasets of human driving  (Fig.~\ref{fig:teaser}, left). Typically, the model is optimized to map raw sensory inputs directly to either low-level control commands (e.g., steering, throttle, braking) \cite{DBLP:journals/corr/abs-1903-10995, DBLP:conf/cvpr/XuGYD17, DBLP:journals/corr/BojarskiTDFFGJM16} or high-level waypoints \cite{DBLP:conf/cvpr/PrakashC021, DBLP:conf/cvpr/HuYCLSZCDLWLJLD23, DBLP:conf/iccv/JiangCXLCZZ0HW23}. In these setups, loss functions are applied solely at the output level. Although effective, this paradigm may provide a weak supervision signal for learning robust and generalizable representations.  
As a result, the model may fail to capture the rich intermediate structures necessary for solving complex driving tasks. 

A promising direction to address this limitation is to introduce additional supervision signals through \emph{deep supervision} --- feature-space regularization applied to intermediate model components. While variants of deep supervision have shown success in other domains \cite{DBLP:conf/aistats/LeeXGZT15, kou2025imacsr}, their application in AD is largely unexplored. Moreover, leveraging GT annotations meaningfully at the feature level, rather than exclusively at the output, is a nontrivial challenge. 

A notable recent advancement in this direction is Vision Language Planning (VLP) \cite{DBLP:conf/cvpr/PanYNMAVR24}, which introduced the use of a frozen text encoder (TE) to regularize the latent space of E2E AD (Fig.~\ref{fig:teaser}, middle). While VLP demonstrates the significant benefits of latent-space guidance with such an encoder, it operates largely in an implicit manner. Building upon this work, our study seeks to provide a formal theoretical perspective on why this training method succeeds, and how it can be generalized to convert available GT data into meaningful latent-space regularization signals. We note that, despite prior framing, the encoder used in VLP (CLIP) is a contrastively-trained VLM rather than a decoder-style large language model (LLM). We adopt the more precise terminology throughout this work.

\begin{figure}[t] 
  \centering
    \includegraphics[width=1.0\linewidth]{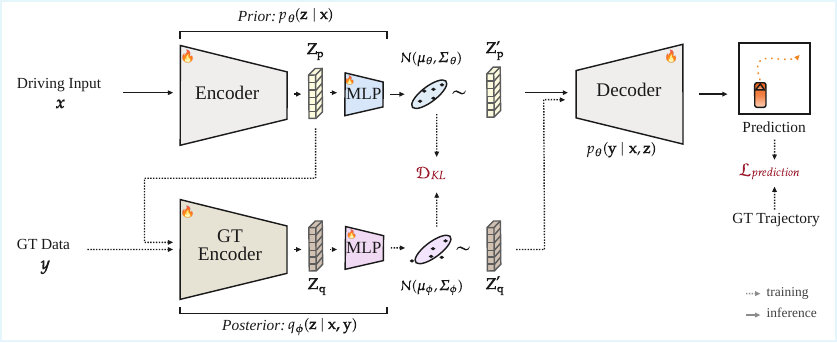}
    \caption{\textbf{Proposed probabilistic deep supervision framework.} PRISM models target latent features as distributions and follows an ELBO-based training objective. GT encoder has an MLP-based architecture, omiting using text encoders.}
  \label{fig:vlp-elbo}
\end{figure}  

Our analysis reveals that VLP's methodology functionally, albeit implicitly, optimizes a simplified version of the Evidence Lower Bound (ELBO), using Dirac distributions as a deterministic approximation of the posterior. Crucially, we demonstrate that VLP's success is driven primarily by this implicit distribution alignment between the prior and posterior (i.e., tying model latents to GT future trajectories), rather than the semantic knowledge encoded by the TE. By formalizing this underlying mechanism, we show how to extract rich regularization signals directly from GT data, greatly streamlining the training process. Our study is conducted on the nuScenes dataset~\cite{DBLP:conf/cvpr/CaesarBLVLXKPBB20} with the VAD-Tiny~\cite{DBLP:conf/iccv/JiangCXLCZZ0HW23} architecture under a shortened training schedule. The main contributions of the work are summarized as follows:

\begin{itemize}
    \item Building upon the success of VLP, we conduct a systematic analysis of key design choices for effective deep supervision in AD, isolating the mechanisms that drive performance.
    \item We introduce a novel probabilistic deep supervision framework for E2E AD  (Fig.~\ref{fig:teaser}, right). In this framework, encoded latents are modeled as reparameterizable distributions to capture uncertainty and are optimized directly via an ELBO objective.
    \item Under this design, our method consistently reduces planning L2 error and collision rate compared to strong E2E baselines on the nuScenes dataset~\cite{DBLP:conf/cvpr/CaesarBLVLXKPBB20} ($\approx 8\%$ lower L2 and $\approx 3\%$ lower collision rate).
\end{itemize}

 \section{RELATED WORK}

\subsection{End-to-End Models}
The advancement of AD has historically diverged into two fundamental paradigms: modular and end-to-end architectures. The \textbf{modular} approach, exemplified by early systems like VaMoRs~\cite{DICKMANNS1987221}, decomposes driving into standalone tasks (e.g., perception, prediction, planning). While this offers interpretability and a parallel development workflow, it suffers from error accumulation and lacks a global optimization objective. Conversely, \textbf{E2E} learning, initially showcased by ALVINN~\cite{DBLP:conf/nips/Pomerleau88} and PilotNet~\cite{DBLP:journals/corr/abs-2010-08776}, maps sensor inputs directly to control signals. This allows the model to learn complex feature interactions but traditionally resulted in "black-box" systems that are difficult to debug and fail to generalize in complex traffic scenarios.

To bridge this gap, modern architectures have evolved toward \textbf{structured E2E} pipelines that implicitly model driving sub-tasks while remaining fully differentiable. BEVFormer~\cite{DBLP:conf/eccv/LiWLXSLQD22} establishes the utility of mapping multi-view images into a unified top-down feature grid, commonly referred to as a bird's-eye view (BEV) representation. Building on this, UniAD~\cite{DBLP:conf/cvpr/HuYCLSZCDLWLJLD23} introduces the first unified, query-driven framework where perception and planning modules are jointly optimized, utilizing learnable queries to facilitate inter-module communication. While constructing dense BEV representations has become a crucial yet computationally demanding task, subsequent works have focused on efficiency and scalability: VAD~\cite{DBLP:conf/iccv/JiangCXLCZZ0HW23} and SparseDrive~\cite{DBLP:journals/corr/abs-2405-19620} replace dense rasterized grids with sparse vectorized representations, drastically reducing computational overhead. Most recently, the field has moved beyond deterministic regression toward generative modeling. GenAD~\cite{DBLP:conf/eccv/ZhengSGZC24} formulates driving as a future scene generation problem, utilizing latent trajectory spaces to capture the multi-modal uncertainty inherent in dynamic environments.
 
\subsection{Deep supervision}
Deep supervision enhances gradient flow, stabilizes optimization, and promotes robust intermediate features by adding auxiliary constraints to intermediate network layers. Originating with pioneering architectures like GoogleNet~\cite{DBLP:conf/cvpr/SzegedyLJSRAEVR15} and DSN~\cite{DBLP:conf/aistats/LeeXGZT15}, this paradigm initially focused on mitigating vanishing gradients. Over time, however, its scope has expanded well beyond traditional supervised scenarios, finding new applications in self-supervised learning~\cite{DBLP:conf/wacv/RenWAZH25} and architecture-agnostic feature regularization~\cite{kou2025imacsr}.

Transitioning to the domain of AD, deep supervision naturally emerges as a mechanism to support interpretable E2E stacks through intermediate tasks. For instance, frameworks like UniAD~\cite{DBLP:conf/cvpr/HuYCLSZCDLWLJLD23} and VAD~\cite{DBLP:conf/iccv/JiangCXLCZZ0HW23} utilize shared BEV representations trained via multiple joint heads. Taking a slightly different path, SparseDrive~\cite{DBLP:journals/corr/abs-2405-19620} applies separate heads to sparse scene representations. Rather than relying on these rule-specific auxiliary tasks, alternative methods like DTCP~\cite{DBLP:journals/corr/abs-2508-18898} explicitly regularize internal features, employing a feature-diversity loss to yield sparser, localized activations that ultimately improve closed-loop performance.

Most recently, a compelling evolution of this concept has emerged: \emph{language-driven} feature regularization, which supervises internal representations using text semantics. A prime example is \textbf{VLP}~\cite{DBLP:conf/cvpr/PanYNMAVR24}, which converts GT annotations into context-rich text and embeds them via a frozen TE to regularize core latent components. This approach achieves consistent improvements in long-tail performance and generalization, adding only minimal training overhead and no inference cost. Building directly on this premise, VLM-AD~\cite{DBLP:journals/corr/abs-2412-14446} elevates the process by leveraging a VLM with structured prompts and auxiliary heads to more effectively distill contextual knowledge into the AD model. Yet, despite the potential of these language-driven methods, identifying theoretically-grounded and optimal ways to leverage GT for latent feature regularization remains an open challenge.

\subsection{Probabilistic Learning and CVAEs}
While standard supervised training relies on Empirical Risk Minimization (ERM) to provide deterministic point estimates~\cite{DBLP:journals/tnn/Vapnik99}, this approach struggles to capture the structured uncertainty inherent in real-world environments. Probabilistic learning frameworks, in contrast, establish a theoretical foundation for learning predictive distributions that capture the spread of likely outcomes, moving beyond the limitations of deterministic residuals. In the context of AD, the future trajectory of a vehicle depends on unobserved factors, such as driver intentions and occluded agents, that underlie the data-generating process. To address this, latent variable models (LVMs) explicitly introduce unobserved variables to encode this hidden structure, allowing them to naturally model aleatoric uncertainty rather than just parameter (epistemic) uncertainty~\cite{bishop2006prml, DBLP:books/lib/Murphy12}.

However, exact inference in classical LVMs is computationally intractable for deep learning. Conditional Variational Autoencoders (CVAEs) resolve this bottleneck through amortized inference~\cite{DBLP:conf/cogsci/GershmanG14}. Instead of optimizing per-sample parameters, CVAEs utilize an inference network with global parameters $\phi$ that conditions on both the input observation $\mathbf{x}$ (e.g., sensor data) and the target $\mathbf{y}$ (e.g., ground-truth future trajectories). This allows the model to approximate the true posterior $p(\mathbf{z} \mid \mathbf{x}, \mathbf{y})$ by maximizing the conditional ELBO.

\begin{figure}[!t]
  \centering
  \includegraphics[width=\linewidth]{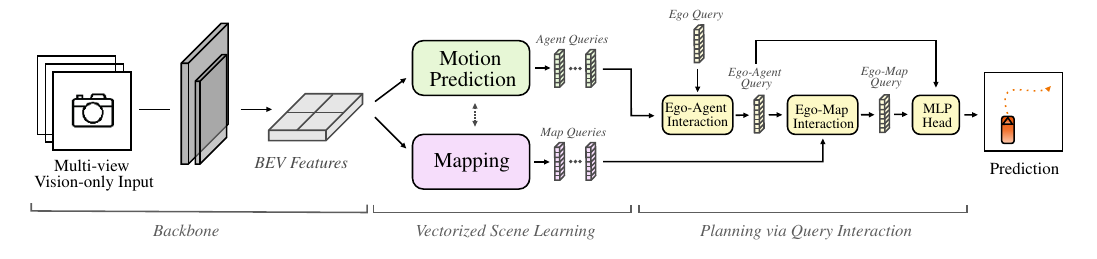}
 \caption{\textbf{VAD \cite{DBLP:conf/iccv/JiangCXLCZZ0HW23} architecture.} The image backbone lifts multi-view inputs into BEV features. A vectorized scene-learning stage then encodes salient scene components (lanes, road boundaries, dynamic agents) as queries. These queries condition an ego-vehicle query via an interaction mechanism, which is subsequently optimized for the planning objective.}
  \label{fig:vad}
\end{figure}

This objective fundamentally balances two goals: reconstruction accuracy (the expected log-likelihood) and regularization toward a learned conditional prior via the Kullback–Leibler (KL) divergence~\cite{kullback1951information}. By leveraging the reparameterization trick~\cite{DBLP:journals/corr/KingmaW13} to backpropagate through stochastic sampling, CVAEs provide a scalable, gradient-based method to align predictive distributions.  We formally adapt this CVAE formulation and its specific parameterization for E2E AD in Section~\ref{sec:method}. 

\section{METHOD}
\label{sec:method}

Building on the empirical successes of VLP~\cite{DBLP:conf/cvpr/PanYNMAVR24}, we aim to formalize latent feature regularization in E2E AD models as a practical training framework for AD pipelines. First, we outline the E2E baseline architecture (VAD~\cite{DBLP:conf/iccv/JiangCXLCZZ0HW23}) and its theoretical learning paradigm. Next, we detail the VLP text-driven supervision paradigm to show how it regularizes the latent space. Finally, we introduce our generalized probabilistic deep supervision framework, transitioning from deterministic feature alignment to probabilistic modeling.

\subsection{VAD: E2E Baseline}
\label{sec:method_vad}

We adopt VAD~\cite{DBLP:conf/iccv/JiangCXLCZZ0HW23} as our base E2E architecture. As illustrated in Fig.~\ref{fig:vad}, VAD processes multi-view images $\mathbf{x}$ to extract BEV features. A vectorized scene-learning stage then encodes salient scene components, such as lanes, road boundaries, and dynamic agents, into distinct queries. These queries interact with a learned ego-vehicle query which is subsequently optimized to output the final planning waypoints $\mathbf{y}$.

Theoretically, VAD fundamentally operates under the paradigm of discriminative learning. It models the conditional distribution $p_{\boldsymbol{\theta}}(\mathbf{y} \mid \mathbf{x})$, where $\boldsymbol{\theta}$ represents the network parameters. The model is optimized via ERM, where the training objective relies purely on the output loss with respect to the GT trajectories. The deterministic baseline optimization objective can be summarized as:
\vspace{-0.2cm}
\begin{equation}
\mathcal{L}_{\mathrm{VAD}} = \mathcal{L}_{\mathrm{ERM}} = \mathbb{E}_{(\mathbf{x}, \mathbf{y}) \sim \mathcal{D}} \left[ -\log p_{\boldsymbol{\theta}}(\mathbf{y} \mid \mathbf{x}) \right].
\label{eq:vad_erm}
\end{equation}

Here, $\mathcal{D}$ represents the distribution of image-trajectory pairs $(\mathbf{x}, \mathbf{y})$. While effective for point-to-point prediction, this approach treats feature distributions as deterministic point estimates and struggles to capture the structured uncertainty inherent in driving environments.

\subsection{VLP: Deterministic Deep Supervision}
\label{sec:method_vlp}

\begin{figure}[!t]
  \centering
  \includegraphics[width=1.0\columnwidth]{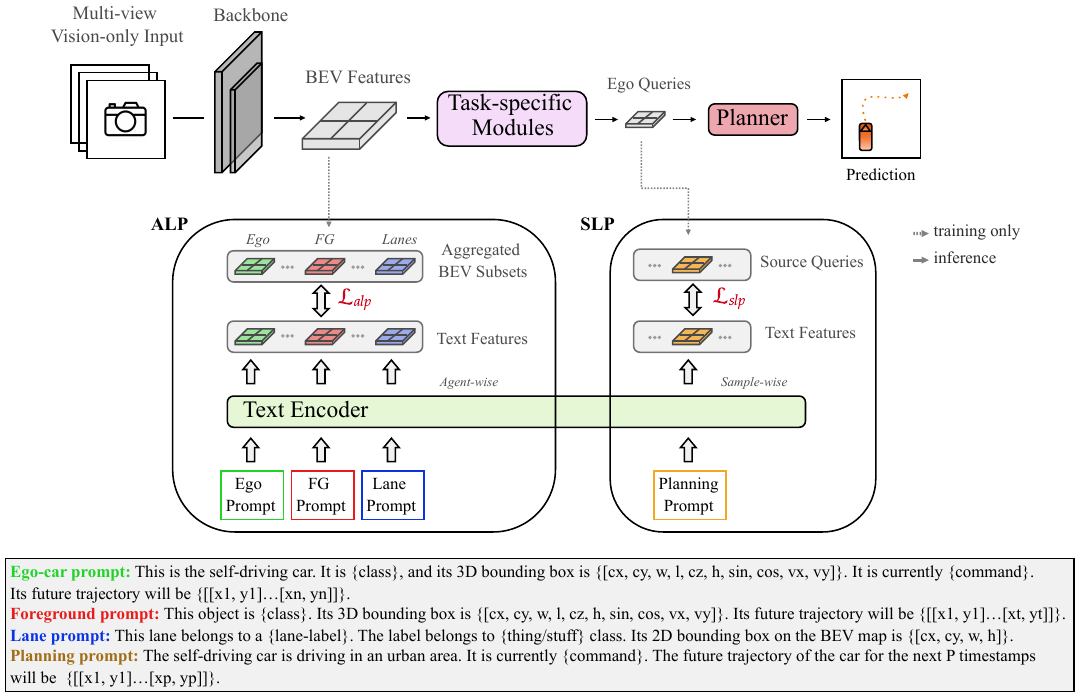}
  \caption{\textbf{Description of VLP \cite{DBLP:conf/cvpr/PanYNMAVR24} framework.} During training, GT annotations are represented as structured prompts, encoded by a frozen text encoder and aligned with internal visual features via a contrastive objective. ALP supervises agent-centric BEV regions, and SLP supervises the ego-vehicle queries used for planning.
 }
  \label{fig:vlp-general-idea}
\end{figure}

To regularize the intermediate representations extracted by VAD, Vision Language Planning (VLP)~\cite{DBLP:conf/cvpr/PanYNMAVR24} introduces a training-only deep supervision paradigm. As shown in Fig.~\ref{fig:vlp-general-idea}, GT scene annotations are converted into natural-language prompts, encoded by a frozen text encoder $\mathbf{E}$, and aligned with internal visual features via a contrastive objective. VLP consists of two components: the \emph{Agent-centric Learning Paradigm (ALP)} for BEV feature regularization, and the \emph{Self-driving-car-centric Learning Paradigm (SLP)} for planning-query regularization. The language branch is discarded at inference, adding no runtime overhead.

\subsubsection{Agent-centric Learning Paradigm (ALP)}
ALP focuses supervision on semantically relevant regions of the BEV representation by grouping agents into three categories: \textit{ego vehicle, foreground objects}, and \textit{lane/map elements} -- $k~\in~\{\mathrm{ego, fg, lane}\}$. Using GT annotations, it extracts local BEV subsets $\mathbf{A}^k_{\mathrm{bev}}$ via 3D bounding box cropping (for  $\mathrm{ego}$ and  $\mathrm{fg}$) or panoptic segmentation (for  $\mathrm{map}$ elements), followed by a pooling operator to standardize spatial dimensions. Simultaneously, target \emph{agent-expectation} features $\mathbf{A}^k_{\mathrm{exp}}$ are generated by passing category-specific text prompts $\mathbf{T}_k$ through the frozen encoder $\mathbf{E}$ and a trainable MLP adapter 
$\mathbf{F}_{\mathrm{bev}}$:
\vspace{-0.3cm}
\begin{align}
    \mathbf{A}^k_{\mathrm{bev}} &= \mathrm{Pool}(\mathrm{Crop/Seg}(\mathrm{BEV}, \mathrm{GT}_k)), \label{eq:alp_bev} \\
    \mathbf{A}^k_{\mathrm{exp}} &= \mathbf{F}_{\mathrm{bev}}(\mathbf{E}(\mathbf{T}_k[\mathbf{y}_k])). \label{eq:alp_exp}
\end{align}
The per-sample features are concatenated into batch-level tensors $\mathbf{A}_{\mathrm{bev}}, \mathbf{A}_{\mathrm{exp}} \in \mathbb{R}^{N_B \times C}$, where $N_B$ is the total number of agents in the batch and $C$ is the channel dimension. ALP employs a symmetric contrastive loss $\mathcal{L}_{\mathrm{alp}}$ to maximize the cosine similarity between corresponding visual-text pairs while minimizing it for all other pairs in the batch, effectively forcing the BEV features to align with the semantic expectations derived from GT data.

Specifically, after applying $\ell_2$ normalization to the features, a pairwise similarity matrix $\mathbf{S}_{\mathrm{pred}}$ is computed and scaled by a learnable logit parameter $\alpha_{\mathrm{alp}}$. Taking the identity matrix $\mathbf{I}_{N_B}$ as the target to enforce one-to-one alignment, the model minimizes a bidirectional cross-entropy loss:
\vspace{-0.2cm}
\begin{equation}
\begin{aligned}
\mathbf{S}_{\mathrm{pred}} &= \alpha_{\mathrm{alp}} (\hat{\mathbf{A}}_{\mathrm{bev}} \hat{\mathbf{A}}_{\mathrm{exp}}^\top), \quad \text{where} \;\; \hat{\mathbf{A}} = \frac{\mathbf{A}}{\lVert \mathbf{A} \rVert_2} \\
\mathcal{L}_{\mathrm{alp}} &= \frac{1}{2} \sum_{d=0}^{1} \mathcal{L}_{\mathrm{CE}}(\mathbf{S}_{\mathrm{pred}}, \mathbf{I}_{N_B}, \mathrm{dim}=d).
\end{aligned}
\label{eq:alp_contrast}
\end{equation}

\subsubsection{Self-driving-car-centric Learning Paradigm (SLP)} 
While ALP stabilizes the global scene representation, SLP shifts the focus to the decision-making bottleneck: the ego-vehicle query. In VAD, ego-vehicle query is constructed as follows. Queries from dedicated modules $\{\mathbf{A}_{\mathrm{query}}\}^{\{k\}}$ are passed through an interaction module $\mathrm{M}_{\mathrm{inter}}$ to aggregate full-scene context into a refined ego-centric feature $\mathbf{E}^{\mathrm{feat}}_{\mathrm{ego}}$. This feature serves as the latent input for the planning decoder $\mathrm{M}_{\mathrm{plan}}$ to predict future waypoints $\mathbf{y}^{\mathrm{plan}}_{\mathrm{pred}}$:
\vspace{-0.1cm}
\begin{equation}
\begin{aligned}
\mathbf{E}^{\mathrm{feat}}_{\mathrm{ego}} &= \mathrm{M}_{\mathrm{inter}}(\mathbf{A}^{\mathrm{ego}}_{\mathrm{query}}, \mathbf{A}^{\mathrm{fg}}_{\mathrm{query}}, \mathbf{A}^{\mathrm{lane}}_{\mathrm{query}}) \in \mathbb{R}^{B \times C}, \\[6pt]
\mathbf{y}^{\mathrm{plan}}_{\mathrm{pred}} &= \mathrm{M}_{\mathrm{plan}}(\mathbf{E}^{\mathrm{feat}}_{\mathrm{ego}}) \in \mathbb{R}^{B \times P \times 2}.
\end{aligned}
\label{eq:slp_fwd}
\end{equation}
Following the contrastive logic of ALP, a planning prompt $\mathbf{E}_{\mathrm{prompt}}$ is constructed from the GT high-level command and the future ego trajectory $\mathbf{y}^{\mathrm{plan}}$. The model is optimized via a bidirectional contrastive loss $\mathcal{L}_{\mathrm{slp}}$ to align the latent ego-query with this language-based expectation. 

Ultimately, the VLP objective modifies the baseline ERM by appending these latent contrastive penalties:
\vspace{-0.2cm}
\begin{equation}
\mathcal{L}_{\mathrm{VLP}} = \mathcal{L}_{\mathrm{ERM}} + \lambda_{\mathrm{alp}} \mathcal{L}_{\mathrm{alp}} + \lambda_{\mathrm{slp}} \mathcal{L}_{\mathrm{slp}}.
\label{eq:vlp_full_loss}
\end{equation}
Crucially, in this formulation, the latent space is regularized without any prior distributional assumptions. Furthermore, the use of a TE to generate $\mathbf{z}_{\mathrm{prompt}}$ is secondary to the overarching theoretical architecture. It primarily serves as a heuristic of choice for deep supervision.
The underlying effectiveness, as we will show, stems from the forced connection between the latent query and the GT future.

\subsection{Probabilistic Deep Supervision Framework}
\label{sec:method_probabilistic}

To provide a theoretically grounded mechanism for latent regularization in this setting and, potentially, for capturing uncertainty inherent in dynamic driving environments, we transition from deterministic point estimates to a principled probabilistic formulation, as depicted in Fig.~\ref{fig:vlp-elbo}. We conceptualize the base AD network as a conditional generative model \mbox{$p_{\boldsymbol{\theta}}(\mathbf{y} \mid \mathbf{x})$, parameterized by $\boldsymbol{\theta}$}. 

We introduce intermediate latent variables $\mathbf{z}$ (representing the regularized ego queries) and model them as Gaussian distributions with diagonal covariance. To properly formalize this as a CVAE, the overall architecture is decomposed into a decoder \mbox{$p_{\boldsymbol{\theta}}(\mathbf{y} \mid \mathbf{x}, \mathbf{z})$}, which generates planning waypoints from sampled latents alongside the input condition, and a conditional prior \mbox{$p_{\boldsymbol{\psi}}(\mathbf{z} \mid \mathbf{x})$}, parameterized by independently learned parameters $\boldsymbol{\psi}$ to model the latent distribution based solely on sensory inputs. Concurrently, GT data $\mathbf{y}$ is encoded into a target approximate posterior distribution \mbox{$q_{\boldsymbol{\phi}}(\mathbf{z} \mid \mathbf{x}, \mathbf{y})$} by an auxiliary inference network parameterized by $\boldsymbol{\phi}$.

Rather than relying on the contrastive loss which lacks distributional assumptions, we optimize the network by maximizing the conditional ELBO objective $\mathcal{J}_{\mathrm{CVAE}}$. This fundamentally aligns the model's predictive distribution with the underlying data-generating process:
\begin{equation}
\begin{aligned}
\mathcal{J}_{\mathrm{CVAE}}(\boldsymbol{\theta}, \boldsymbol{\phi}, \boldsymbol{\psi}) &= \mathbb{E}_{\mathbf{z} \sim q_{\boldsymbol{\phi}}(\mathbf{z} \mid \mathbf{x}, \mathbf{y})} \!\left[ \log p_{\boldsymbol{\theta}}(\mathbf{y} \mid \mathbf{x}, \, \mathbf{z}) \right] \\
&\quad - \beta \, D_\mathrm{KL}\!\left( q_{\boldsymbol{\phi}}(\mathbf{z} \mid \mathbf{x}, \mathbf{y}) \,\|\, p_{\boldsymbol{\psi}}(\mathbf{z} \mid \mathbf{x}) \right).
\end{aligned}
\label{eq:elbo}
\end{equation}

During training, latents are sampled from $q_{\boldsymbol{\phi}}$ via the reparameterization trick~\cite{DBLP:journals/corr/KingmaW13} to compute the expected reconstruction loss, while the $\beta$-scaled KL divergence regularizes the conditional prior $p_{\boldsymbol{\psi}}$ toward the GT-informed posterior. Crucially, at inference, the posterior encoder is discarded. Latents are instead sampled from the prior $p_{\boldsymbol{\psi}}$, adding zero computational overhead to the baseline E2E model.

Ultimately, our core architectural contribution is two-fold: replacing a deterministic contrastive loss with a principled probabilistic framework, and shifting from a reliance on text encoders to an efficient, MLP-based encoding. Further implementation specifics and detailed architectural parameterizations are provided in Sec.~\ref{sec:experiments}, with additional encoding comparisons and network details documented in Appendix A and Appendix B.

\section{EXPERIMENTS}
\label{sec:experiments}

\begin{table*}[!t]
\caption{\textbf{Quantitative Evaluation of the Probabilistic Framework.}}
\vspace{-0.2cm}
\label{tab:main_results}
\centering
\small
\setlength{\tabcolsep}{3pt}
\renewcommand{\arraystretch}{1.05}
\begin{tabular}{@{}l | ccc | ccc@{}}
\toprule
\multirow{2}{*}{\textbf{Model / Configuration}} &
\multicolumn{3}{c|}{\textbf{L2 (m) $\downarrow$}} &
\multicolumn{3}{c}{\textbf{Col. Rate (\%) $\downarrow$}} \\
& \textbf{1s} & \textbf{2s} & \textbf{3s} & \textbf{1s} & \textbf{2s} & \textbf{3s} \\
\midrule
VAD~\cite{DBLP:conf/iccv/JiangCXLCZZ0HW23} & $0.44_{\pm 0.03}$ & $0.77_{\pm 0.06}$ & $1.17_{\pm 0.10}$ & $0.74_{\pm 0.13}$ & $0.84_{\pm 0.24}$ & $1.10_{\pm 0.29}$ \\
\addlinespace[3pt]
\multicolumn{7}{@{}>{\columncolor{gray!12}}l@{}}{\textit{Probabilistic Framework (VAD-VLP, SLP-Only)}} \\
\addlinespace[1pt]
\quad Deterministic SLP  & $0.33_{\pm 0.01}$ & $0.57_{\pm 0.01}$ & $0.88_{\pm 0.02}$ & $0.21_{\pm 0.13}$ & $0.29_{\pm 0.13}$ & $0.46_{\pm 0.13}$ \\
\quad Probabilistic ELBO ($S=1$) & $0.37_{\pm 0.03}$ & $0.63_{\pm 0.04}$ & $0.95_{\pm 0.05}$ & $0.28_{\pm 0.03}$ & $0.27_{\pm 0.02}$ & $\mathbf{0.42}_{\pm 0.01}$ \\
\quad \textbf{Probabilistic ELBO} ($S=2$) & $\mathbf{0.29}_{\pm 0.01}$ & $\mathbf{0.52}_{\pm 0.01}$ & $\mathbf{0.84}_{\pm 0.02}$ & $\mathbf{0.13}_{\pm 0.02}$ & $\mathbf{0.26}_{\pm 0.02}$ & $0.66_{\pm 0.13}$ \\
\bottomrule
\end{tabular}
\begin{minipage}{0.85\linewidth}
\vspace{0.5em}
{\normalfont \footnotesize Values report the mean $\pm$ standard deviation of models evaluated at epoch 35. The VAD baseline is averaged over three runs. Our ELBO-based probabilistic formulation ($S=2$ samples) outperforms the optimized deterministic SLP baseline, achieving the strongest overall planning performance and the lowest L2 errors across all temporal horizons.\par}
\end{minipage}
\end{table*}

\textbf{Implementation Details.}
We evaluate our framework on the nuScenes dataset~\cite{DBLP:conf/cvpr/CaesarBLVLXKPBB20} using the \texttt{VAD-Tiny} configuration. All models are trained with a global batch size of 8 distributed across 2$\times$NVIDIA A100 GPUs. Given the substantial computational cost (60 epochs require $\approx 12$ days), we limit training to 35 epochs. Intermediate validation confirms that performance stabilizes sufficiently at this stage to reliably evaluate design variations. To stabilize the probabilistic training, we apply gradient isolation, cutting gradients flowing to the prior features from the GT-encoder branch, and use a KL weighting factor of $\beta = 0.1$. We report L2 trajectory error (m) and Collision Rate (\%) across 1s, 2s, and 3s planning horizons. Our codebase builds upon the official VAD implementation~\footnote{\footnotesize\url{https://github.com/hustvl/VAD}}. As the official VLP codebase is not publicly available, we utilized a partial release provided by the authors~\cite{DBLP:conf/cvpr/PanYNMAVR24}. However, because this version did not support direct reproduction of the reported results, we performed extensive refactoring and added missing components prior to conducting our experiments. Our re-implementation successfully reproduces the reported L2 planning performance, though we observed a performance gap in Collision Rate compared to the VLP baseline. This appears to be a known issue with an underlying VAD codebase, reported by other users. 
\subsection{Baseline Reproduction and Training Dynamics}

\begin{figure}[!htbp]
  \centering
  \includegraphics[width=\linewidth]{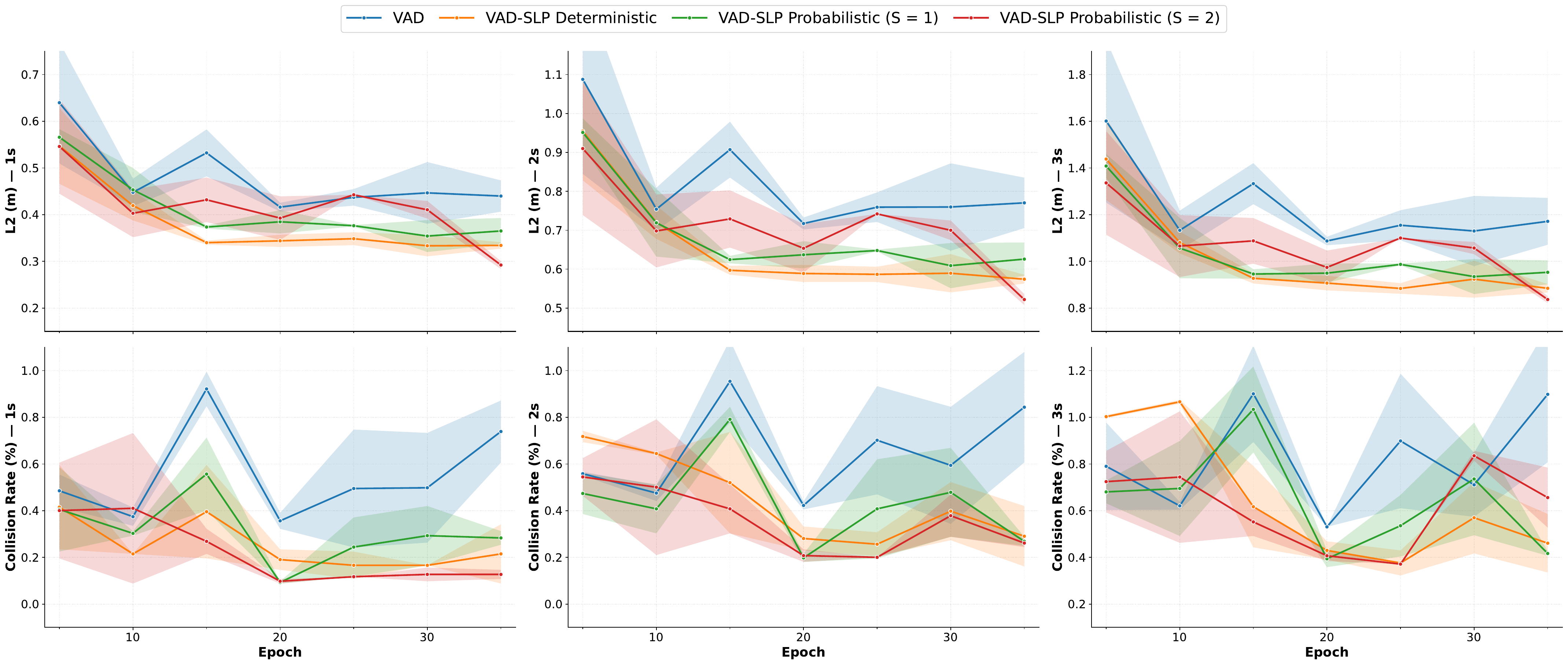}
 \caption{\textbf{Learning curves comparing original deterministic VLP and our probabilistic extension over 35 epochs.} Thick lines denote the mean over two independent runs, with shaded regions indicating $\pm1$ standard deviation. We use only the SLP branchs. $S$ denotes the number of samples used to evaluate the reconstruction loss term. The probabilistic approach matches deterministic baseline performance and yields modest gains when the ELBO reconstruction term is optimized with $S = 2$ samples from latents.}
  \label{fig:vlp-slp-prob-main}
\end{figure} 

Training E2E AD models is inherently complex, computationally intensive, and highly stochastic. Nevertheless, results in the literature are commonly reported as single-point estimates~\cite{DBLP:conf/eccv/LiWLXSLQD22, DBLP:conf/iccv/JiangCXLCZZ0HW23, DBLP:journals/corr/abs-2405-19620, DBLP:conf/cvpr/PanYNMAVR24}, without details on result selection, convergence behavior, or performance variability. The absence of such information limits our understanding of the training dynamics, leaving the community vulnerable to selective reporting and making the distinction between genuine progress and random variance increasingly opaque.

To address this, we systematically reproduce the VAD and VLP baselines and actively report learning curves and uncertainty estimates throughout this work. As shown in Fig.~\ref{fig:vlp-slp-prob-main}, visualizing the mean and standard deviation over multiple independent runs reveals significant variance, particularly in collision rates. By explicitly reporting these curves, we provide a more transparent and rigorous foundation for analyzing architectural modifications, representing a critical methodological step for proper model analysis. Due to the high variance in observed collision rates, we rely primarily on the L2 error for robust model comparison.

\subsection{Iterative Architectural Search}
To build a generalized and highly efficient framework, we first conduct a systematic ablation study on the deterministic VLP baseline. Our goal is to isolate the true drivers of its performance, eliminate redundant complexities, and establish a streamlined foundation for our probabilistic extension.

\subsubsection{Isolating Spatial and Query-Level Supervision (ALP vs. SLP)}
In the original formulation, deep supervision is applied both to semantically relevant regions of the BEV representation (ALP) and to the interaction queries prior to the planning stage (SLP). As detailed in Table~\ref{tab:alp-vs-slp-ablation}, adding ALP to SLP yields only marginal performance improvements while significantly increasing architectural complexity, specifically regarding the aggregation of salient BEV regions and the handling of variable numbers of scene elements. Furthermore, SLP-only training exhibits smoother convergence and lower variance across runs. Given the increased computational overhead and less stable convergence of the combined approach, we omit ALP alignment in favor of SLP-only supervision, streamlining the architecture with no substantial loss in performance.
\begin{table}[!htbp]
\caption{\textbf{ALP vs. SLP Ablation.}}
\vspace{-0.2cm}
\label{tab:alp-vs-slp-ablation}
\centering
\small
\setlength{\tabcolsep}{3pt}
\renewcommand{\arraystretch}{1.05}
\resizebox{\columnwidth}{!}{%
\begin{tabular}{@{}l | ccc | ccc@{}}
\toprule
\multirow{2}{*}{\textbf{Model / Configuration}} &
\multicolumn{3}{c|}{\textbf{L2 (m) $\downarrow$}} &
\multicolumn{3}{c}{\textbf{Col. Rate (\%) $\downarrow$}} \\
& \textbf{1s} & \textbf{2s} & \textbf{3s} & \textbf{1s} & \textbf{2s} & \textbf{3s} \\
\midrule
VAD & $0.44_{\pm 0.03}$ & $0.77_{\pm 0.06}$ & $1.17_{\pm 0.10}$ & $0.74_{\pm 0.13}$ & $0.84_{\pm 0.24}$ & $1.10_{\pm 0.29}$ \\
\addlinespace[3pt]
\multicolumn{7}{@{}>{\columncolor{gray!12}}l@{}}{\textit{VAD–VLP}} \\
\addlinespace[1pt]
\quad ALP & $0.31_{\pm 0.02}$ & $0.54_{\pm 0.03}$ & $0.84_{\pm 0.05}$ & $0.49_{\pm 0.21}$ & $0.53_{\pm 0.25}$ & $0.71_{\pm 0.30}$ \\
\quad SLP & $0.32_{\pm 0.02}$ & $0.56_{\pm 0.02}$ & $0.87_{\pm 0.02}$ & $0.19_{\pm 0.11}$ & $0.27_{\pm 0.11}$ & $0.44_{\pm 0.11}$ \\
\quad ALP + SLP & $0.31_{\pm 0.03}$ & $0.55_{\pm 0.05}$ & $0.87_{\pm 0.06}$ & $0.40_{\pm 0.13}$ & $0.35_{\pm 0.11}$ & $0.43_{\pm 0.09}$ \\
\bottomrule
\end{tabular}%
}
\begin{minipage}{\linewidth}
\vspace{0.5em}
{\normalfont \footnotesize Entries denote the mean $\pm$ standard deviation over three independent runs (35 epochs). SLP-only training exhibits smoother convergence and lower collision rates. Adding ALP yields only marginal improvements while introducing additional complexity from salient BEV region aggregation.\par}
\end{minipage}
\end{table}

\subsubsection{The Role of the Text Encoder in GT Encoding}
The baseline framework heavily relies on a frozen TE (CLIP) to encode GT scene annotations into structured textual prompts. To evaluate whether these linguistic priors are strictly necessary, we replace the TE and its adapter with a simple, fully trainable MLP-based encoder that directly processes GT attributes (e.g., ego trajectories and driving commands). 

As shown in Table~\ref{tab:vlp-encoder}, the fully MLP-based encoder matches or slightly surpasses the CLIP-based configuration. This strongly suggests that the performance gains stem from the regularizing effect of the GT injection itself, rather than from the abstract linguistic representations. Consequently, we replace the frozen TE with this MLP-based encoding, further reducing computational overhead.
\begin{table}[!htbp]
\caption{\textbf{GT Encoder Ablation.}}
\vspace{-0.2cm}
\label{tab:vlp-encoder}
\centering
\small
\setlength{\tabcolsep}{3pt}
\renewcommand{\arraystretch}{1.05}
\resizebox{\columnwidth}{!}{%
\begin{tabular}{@{}l | ccc | ccc@{}}
\toprule
\multirow{2}{*}{\textbf{Model / Configuration}} &
\multicolumn{3}{c|}{\textbf{L2 (m) $\downarrow$}} &
\multicolumn{3}{c}{\textbf{Col. Rate (\%) $\downarrow$}} \\
& \textbf{1s} & \textbf{2s} & \textbf{3s} & \textbf{1s} & \textbf{2s} & \textbf{3s} \\
\midrule
VAD & $0.44_{\pm 0.03}$ & $0.77_{\pm 0.06}$ & $1.17_{\pm 0.10}$ & $0.74_{\pm 0.13}$ & $0.84_{\pm 0.24}$ & $1.10_{\pm 0.29}$ \\
\addlinespace[3pt]
\multicolumn{7}{@{}>{\columncolor{gray!12}}l@{}}{\textit{VAD–VLP}} \\
\addlinespace[1pt]
\quad CLIP text encoder & $0.31_{\pm 0.03}$ & $0.55_{\pm 0.05}$ & $0.87_{\pm 0.06}$ & $0.40_{\pm 0.13}$ & $0.35_{\pm 0.11}$ & $0.43_{\pm 0.09}$ \\
\quad MLP encoder & $0.29_{\pm 0.02}$ & $0.53_{\pm 0.03}$ & $0.84_{\pm 0.03}$ & $0.28_{\pm 0.25}$ & $0.35_{\pm 0.24}$ & $0.50_{\pm 0.18}$ \\
\bottomrule
\end{tabular}%
}
\begin{minipage}{\linewidth}
\vspace{0.5em}
{\normalfont \footnotesize Entries report mean $\pm$ standard deviation over three runs (35 training epochs). The fully MLP-based encoder matches, and slightly improves on, the CLIP-based setup, suggesting the frozen text encoder is not the primary source of VLP gains.\par}
\end{minipage}
\end{table}
\subsubsection{Alignment Objectives and Temporal Injection Horizons}
Beyond structural components, we ablate the optimization objective and the temporal horizons of the injected GT signals. First, while the baseline utilizes a symmetric contrastive objective for feature alignment, our experiments show that a standard Mean Squared Error (MSE) loss serves as a highly competitive and simpler alternative, with the contrastive loss providing only minor improvements. Second, we investigate the temporal bounds of the injected ego trajectory. Providing the model with past trajectories yields a negligible learning signal. Conversely, injecting future ego trajectories provides the largest performance gains. Extending this future trajectory to longer temporal horizons offers additional, but strictly marginal, improvements.

\subsection{Proposed Architecture and Quantitative Comparison}

We utilize the simplified SLP-only configuration, paired with a fully MLP-based encoder and future-trajectory GT injection, as the efficient foundation for our probabilistic framework. Building on the mechanistic insights from our previous ablations, we evaluate our principled probabilistic extension against both the base VAD architecture and deterministic VLP baseline (Table~\ref{tab:main_results}).

We observe that aggregating multiple posterior samples for the ELBO reconstruction term is critical for stabilization and performance. Drawing multiple samples is a well-established technique in VAE literature to tighten the ELBO bound and reduce estimator variance. In our framework, it adds only a negligible computational overhead. As detailed in Table~\ref{tab:main_results} and Fig.~\ref{fig:vlp-slp-prob-main}, when drawing $S=2$ samples during training, the probabilistic framework consistently outperforms the deterministic VLP baseline. Across the planning horizons, this yields an $\approx 8\%$ average reduction in L2 error and achieves the lowest collision rates at the 1s and 2s horizons.

This performance gain is theoretically well-grounded. Our analysis suggests that the original VLP framework can be viewed as an implicit ELBO optimization that utilizes a degenerate Dirac distribution as its variational posterior. By contrast, our framework adopts a broader family of approximating distributions —
specifically, a reparameterizable Gaussian. By transitioning from a point-estimate to a distributional posterior, we significantly tighten the gap between the variational approximation and the true posterior. While both remain approximations, our approach provides an additional expressivity intended to capture the complex alignments between latent features and GT trajectories.

\subsection{Analysis of the Learned Generative Model}

Despite the quantitative improvements in planning accuracy, an analysis of the generative behavior revealed a collapse toward uni-modal distributions. When multiple samples were drawn from the prior at inference time, the resulting trajectories were nearly identical, indicating the network failed to learn the full multi-modal diversity of driving maneuvers. In its current form, we therefore understand the main practical benefit of the probabilistic formulation to be stronger latent regularization and improved planning accuracy, rather than richer multi-modal generation. Its potential for uncertainty modeling remains to be demonstrated.

We attribute this lack of diversity to two primary factors. First, the structural constraints of the underlying VAD architecture inherently limit expressivity. VAD treats trajectory generation as a deterministic selection over independent heads rather than a continuous, multi-modal predictive space. Second, while the simple MLP-based posterior encoder provides a theoretically grounded target, it may lack the depth to capture the high-dimensional interactions between dynamic agents. Crucially, however, this limitation is not unique to our approach. Our earlier analysis suggests that the far more complex, frozen CLIP text encoders in VLP likewise serve as high-quality point-estimates rather than diverse generative priors. By replacing the "black-box" text-encoder backbone with a transparent MLP, we achieve superior planning performance while making these latent dynamics, such as mode collapse, explicitly measurable.

Ultimately, these findings offer a vital perspective on the current state-of-the-art, within the scope of the single dataset and architecture studied here. Our ELBO-based formulation moves the field beyond heuristic contrastive losses toward a setup where model behavior is theoretically interpretable. Even in the presence of mode collapse, the framework serves as a more powerful regularizer for planning accuracy than its deterministic counterparts. It establishes a foundation for future work, providing both a high-performance baseline and a principled diagnostic tool for the next generation of truly generative driving architectures.

\section{CONCLUSION AND DISCUSSION}
\label{sec:conclusion}

This work systematically demystifies deep supervision for E2E AD and establishes a formal probabilistic foundation for latent feature regularization. Through rigorous ablations, we demonstrated that the benefits of recent methods like VLP arise not from the reasoning capabilities of the complex text encoders they employ, but from some form of implicit deterministic alignment between model latents and GT distributions. This suggests that much of the current state-of-the-art relies on effective point-estimate regularization rather than the semantic priors often credited.

Leveraging these insights, we introduced a probabilistic framework that replaces heuristic contrastive alignment with a principled Variational Inference objective. Specifically, we model latent representations as reparameterizable distributions trained via the ELBO.  Evaluated on the nuScenes dataset, our framework consistently outperforms competitive baselines in planning accuracy while incurring zero inference overhead.

Within the scope of this work, the results indicate that incorporating structured GT signals into latent spaces is a promising and theoretically grounded blueprint for training lower-error AD models. To address the observed uni-modal collapse, future work can focus on integrating this ELBO-based training paradigm with inherently multi-modal, generative E2E architectures (e.g., GenAD~\cite{DBLP:conf/eccv/ZhengSGZC24}). Furthermore, improving the expressivity of the posterior encoder, such as through attention mechanisms, could prevent early KL collapse and capture richer agent-map interactions. Finally, expanding this generalized framework to incorporate diverse GT modalities presents an exciting avenue for developing scalable, interpretable, and safe AD systems.

\bibliographystyle{IEEEtran}
\bibliography{IEEEabrv,bibliography}

\appendix

\section*{A. GT Data Encoders}

\label{sec:appendix-gt-data-encoders}
\begin{figure}[!htbp]
  \centering
  \subfloat[]{
    \includegraphics[width=0.4\linewidth]{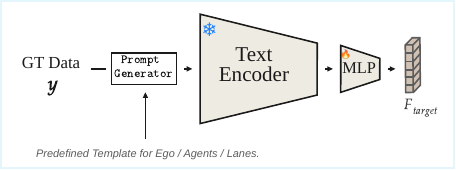}
    \label{fig:vlp-enc-llm}
  }
  \vspace{-1em}
  \subfloat[]{
    \includegraphics[width=1.0\linewidth]{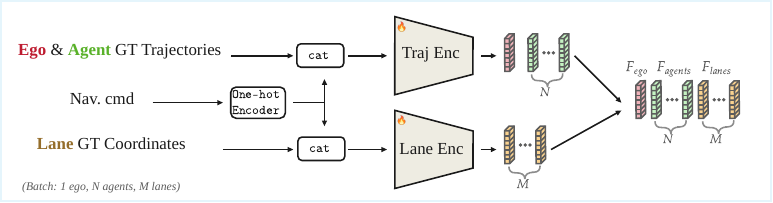}
    \label{fig:vlp-enc-mlp-alp}
  }
  \vspace{-1em}
  \subfloat[]{
    \includegraphics[width=0.6\linewidth]{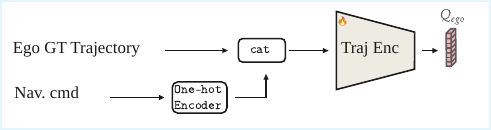}
    \label{fig:vlp-enc-mlp-slp}
  }
  
  \caption{Overview of GT data encoders that transform input data into target features for regularization. (a) Text-based encoder employed in the original VLP framework \cite{DBLP:conf/cvpr/PanYNMAVR24}. (b) and (c) MLP-based encoders proposed in this work for the ALP and SLP components of VLP, respectively.}
  \label{fig:vlp-enc-overview}
\end{figure}
This section describes GT data encoders that map input annotations to target features for deep supervision within the VLP framework \cite{DBLP:conf/cvpr/PanYNMAVR24} (Fig.~\ref{fig:vlp-enc-overview}). We discuss the original text-based encoder (TE) and our proposed MLP-based alternative. 
\subsubsection{Text-based Encoder} 
Fig.~\ref{fig:vlp-enc-llm} illustrates the TE encoder used as the default in VLP. GT data are first formatted into text prompts following a predefined template that differs between ALP and SLP (an example is shown in Fig.~\ref{fig:vlp-general-idea}). The prompt is processed by a frozen TE (\texttt{CLIP/RN50x64}), after which two trainable MLP layers project the text features to the required shape. In ALP, each scene agent (including the ego) and each lane instance are encoded into separate feature vectors. In SLP, the same mechanism is used to produce \texttt{ego-agent} and \texttt{ego-map} interaction queries.

\subsubsection{MLP encoder}
Figs.~\ref{fig:vlp-enc-mlp-alp} and \ref{fig:vlp-enc-mlp-slp} show our MLP-based encoders for ALP and SLP, respectively. In the ALP encoder, GT agents and lanes are transformed into per-element feature vectors. Ego and agent trajectories are concatenated with a one-hot navigation command and then passed through a trajectory encoder consisting of a small stack of trainable MLP layers. To semantically distinguish trajectories from lane geometry, lane coordinates are processed by a separate MLP prior to concatenation with the navigation command. In the simplest configuration, we use two layers for both encoders. Notably, we remove the frozen TE and omit 3D bounding-box attributes for agents, which we empirically found to be non-critical in this setting. SLP follows the same principle, combining the ego trajectory with the one-hot navigation command via a compact trajectory encoder.

\section*{B. Detailed Network Architectures}
\label{sec:appendix-details}

Figure~\ref{fig:vlp-elbo} summarizes the probabilistic reformulation. In the original deterministic setup, SLP regularizes both the \texttt{ego-agent} and \texttt{ego-map} interaction queries. However, in the \texttt{VAD} pipeline the \texttt{ego-map} query is predicted from the \texttt{ego-agent} query. To avoid redundant supervision and reduce coupling, we regularize only the \texttt{ego-agent} query.

\subsubsection{Prior and posterior queries}
We denote the query produced by the original encoders as the \texttt{prior\_query}. To construct a distributional target, we generate a corresponding \texttt{posterior\_query} from GT annotations while conditioning on the \texttt{prior\_query}. The \texttt{posterior\_encoder} is a fully MLP-based module (conceptually related to Fig.~\ref{fig:vlp-enc-mlp-slp}):

\begin{itemize}
    \item \textbf{Embedding heads (prior branch).} The \texttt{prior\_query} is processed by two independent MLP heads to produce features used for the mean and covariance paths, respectively. Each head consists of linear layers with hidden dimensions \texttt{[D, 2D, D]}.
    \item \textbf{GT path.} The future GT trajectory, concatenated with a one-hot driving command, is passed through two identical MLP heads (for mean and covariance paths), each with linear layers \texttt{[D/2, D, 2D, 2D, D, D]}.
    \item \textbf{Feature fusion and projection.} Corresponding features from the prior and GT branches (mean-to-mean, covariance-to-covariance) are concatenated and projected by two independent MLPs to produce the final parameters. The projection stack uses \texttt{Linear} layers \texttt{[2D, 2D{+}D/2, 2D{+}D/2, 2D, D, D]}.
    \item \textbf{Stable covariance parameterization.} We use a diagonal covariance parameterization, predicting per-dimension log-variance and computing the standard deviation as \texttt{std = torch.exp(0.5 * log\_var)}. The diagonal entries are then assembled into a lower-triangular scale matrix to instantiate \texttt{torch.distributions.MultivariateNormal}.

\end{itemize}

\subsubsection{Source (prior) distribution}
For the \texttt{prior\_query} $\rightarrow$ \texttt{prior} distribution (our source distribution in the ELBO), we employ two independent MLP heads — for the mean and for the covariance. Each head is a stack of linear layers with dimensions \texttt{[D, 2D, 2D, D]}.

\subsubsection{Implementation notes}
We implement all modules in \texttt{PyTorch}, using linear layers with \texttt{ReLU} activations, and \texttt{LayerNorm}. Hidden dimension is set to $\texttt{D} = 256$.






\end{document}